\documentclass[referee,sn-basic]{sn-jnl}

\usepackage{soul}  
\usepackage{makecell}  
\usepackage{float}  
\restylefloat{table}
\usepackage{caption}  
\usepackage{subcaption}  
\usepackage[table]{xcolor}  
\usepackage{orcidlink}  

\usepackage{graphicx}%
\usepackage{multirow}%
\usepackage{amsmath,amssymb,amsfonts}%
\usepackage{amsthm}%
\usepackage{mathrsfs}%
\usepackage[title]{appendix}%
\usepackage{xcolor}%
\usepackage{textcomp}%
\usepackage{manyfoot}%
\usepackage{booktabs}%
\usepackage{algorithm}%
\usepackage{algorithmicx}%
\usepackage{algpseudocode}%
\usepackage{listings}%

\begin{document}

\title[Article Title]{The Impact of Variable Ordering on Bayesian Network Structure Learning}


\author*[1]{\fnm{Neville K.}
            \sur{Kitson}
            \orcidlink{0000-0002-7970-1453}}
\email{n.k.kitson@qmul.ac.uk}

\author[1]{\fnm{Anthony C.} 
           \sur{Constantinou}
           \orcidlink{0000-0001-7147-6821}}
\email{a.constantinou@qmul.ac.uk}

\affil[1]{\orgdiv{Machine Intelligence and Decision Systems (MInDS) group},
\orgname{Queen Mary University of London}, \orgaddress{\street{Mile End Road}, \city{London}, \postcode{E1 4NS}, \country{United Kingdom}}}


\abstract{Causal Bayesian Networks (CBNs) provide an important tool for reasoning under uncertainty with potential application to many complex causal systems. Structure learning algorithms that can tell us something about the causal structure of these systems are becoming increasingly important. In the literature, the validity of these algorithms is often tested for sensitivity over varying sample sizes, hyper-parameters, and occasionally objective functions, but the effect of the order in which the variables are read from data is rarely quantified. We show that many commonly-used algorithms, both established and state-of-the-art, are more sensitive to variable ordering than these other factors when learning CBNs from discrete variables. This effect is strongest in hill-climbing and its variants where we explain how it arises, but extends to hybrid, and to a lesser-extent, constraint-based algorithms. Because the variable ordering is arbitrary, any significant effect it has on learnt graph accuracy is concerning, and raises questions about the validity of both many older and more recent results produced by these algorithms in practical applications and their rankings in performance evaluations.}

\keywords{Bayesian Networks,  Directed Acyclic Graphs, probabilistic graphical models, structure learning}



\maketitle

\section{Introduction}
\label{sect:introduction}

\subsection{Bayesian Networks}

Bayesian Networks (BNs) are probabilistic graphical models that compactly represent the independence and dependence relationships between variables \citep{koller2009probabilistic,darwiche2009modeling}. These relationships are encapsulated in a Directed Acyclic Graph (DAG) where each node represents a variable of interest, and the edges represent dependencies between two variables. The form and strength of these dependencies are quantified by the model parameters. The DAG obeys the Local Markov property which states that a variable is conditionally independent of all its non-descendants in the graph given its parents. The DAG and model parameters define the global probability distribution of the variables, which, because of the Local Markov property, can be defined succinctly as

\begin{equation}
P(X_1, X_2, ..., X_n)=\prod_{i=1}^n P(X_i | \textbf{Pa}(X_i))
\end{equation}
where $X_1, X_2, ..., X_n$ are the variables in the model, and $\textbf{Pa}(X_i)$ are the parents of $X_i$. Also flowing from the Local Markov property, \textit{d-separation} \citep{pearl1988probabilistic} is a graphical property of the DAG that can be used to determine the conditional independence relationships implied by the DAG. Another key benefit of a BN is that it can be used to answer \textit{probabilistic queries}, that is, compute the marginal unconditional or conditional probability distribution for any subset of variables.

\begin{figure}[htp]
    \centering
    \includegraphics[width=12cm]{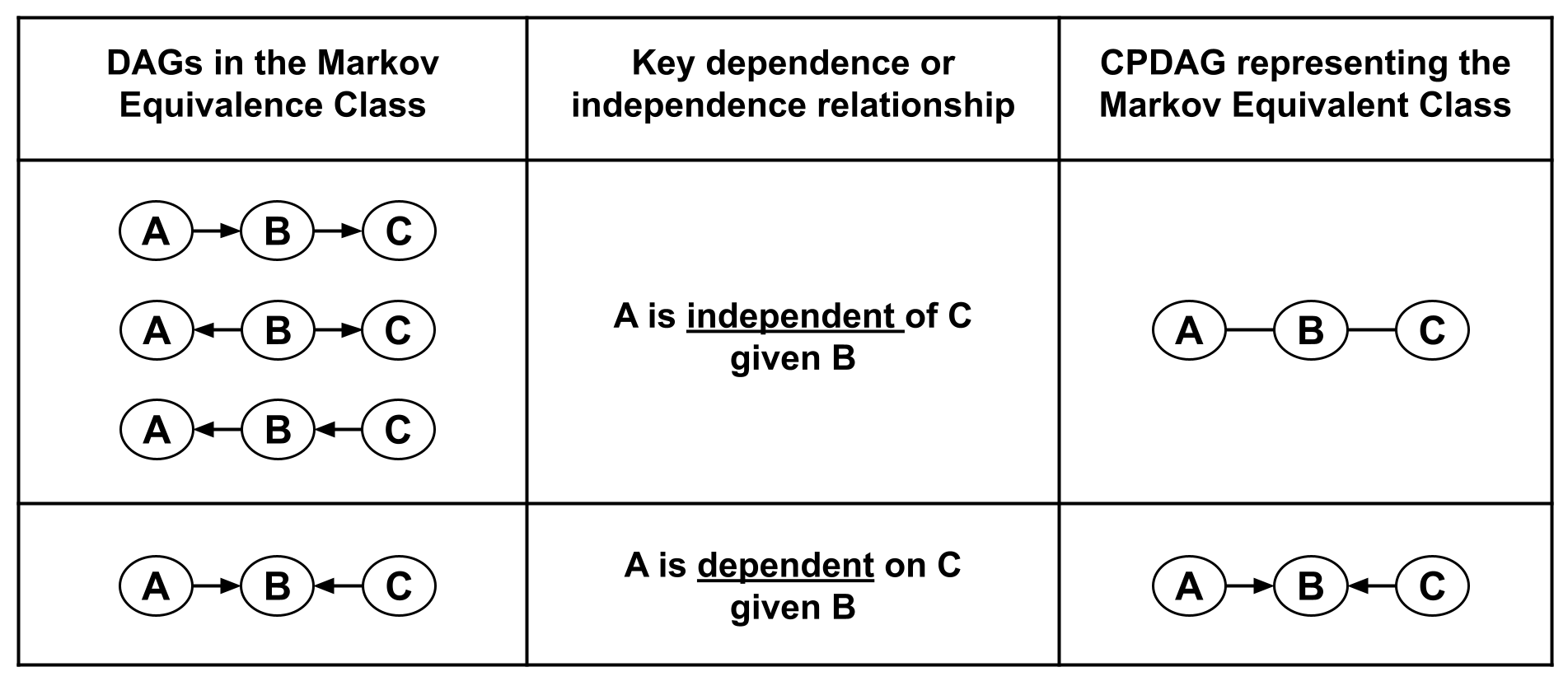}
    \caption{DAGs with three variables and two edges, their distinctive independence or dependence relationships, the MECs they belong to and the CPDAG representing the MEC.}
    \label{fig:ord_mec_examples}
\end{figure}

In general, more than one DAG can represent the same set of independence and dependence relationships and these DAGs are said to be \textit{Markov Equivalent}. The set of DAGs which are Markov Equivalent to one another is termed the Markov Equivalent Class (MEC) \citep{verma1990equivalence}. The MEC is often represented by a Completed Partially Directed Acyclic Graph (CPDAG) where undirected edges, termed \textit{reversible}, indicate edges that have different orientations amongst the members of the MEC. Figure \ref{fig:ord_mec_examples} shows some example MECs and their member DAGs with three variables and two edges. A DAG can be transformed to any other Markov Equivalent DAG through a sequence of \textit{covered arc} reversals. A covered arc is one where both nodes have the same parents ignoring the nodes connected by the edge itself, that is $X_i \longrightarrow X_j$ is a covered arc if:
\begin{equation} \label{eqn:covered}
\textbf{Pa}(X_i)) = \textbf{Pa}(X_j)) \setminus \{ X_i \}
\end{equation}

\subsection{Causal Bayesian Networks}

A Causal Bayesian Network (CBN) is a subclass of BNs where we further assume that each directed edge represents a causal relationship in the real world, so that the edge orientation $A \longrightarrow B$ asserts that $A$ is the cause of $B$ \citep{spirtes2000causation,pearl2009causality}. Our understanding of causality in the real world implies that if we intervene on variable $A$ to change its value, then the value of $B$ will change and that, the change in $A$ will temporally precede that of $B$. Analogously to a general BN, a CBN makes the \textit{Causal Markov Assumption}, that a variable is independent of its non-effects given its causes.

A CBN allows us to model a perfect intervention on a variable $A$ whereby we 'disconnect' $A$ from its usual causes and instead set its value. The modified DAG is usually referred to as a \textit{manipulated} or \textit{mutilated} DAG. We then observe the new probability distribution of other variables, say $B$, which is denoted by $P(B \, | \, do(A))$. Note that, in general, $P(B \, | \, A) \ne P(B \, | \, do(A))$. A CBN can thus answer \textit{interventional queries}; that is, model the effects of interventions in the real world.

As we discussed previously, we cannot generally orientate all edges from observational data, we can usually only learn a CPDAG containing some undirected edges. This is not a problem if we only wish to perform probabilistic queries, since any DAG in the MEC will produce the same results. However, the different DAGs in a MEC will generally produce different results from interventional queries, and it is therefore important to try and orientate as many arcs as possible in a CBN. Undirected edges in the CPDAG can be orientated by making use of interventional data \citep{koller2009probabilistic}. This is possible because the data distributions in the manipulated DAGs can provide extra information which further orientates some edges.

\subsection{Learning Bayesian Networks}

The key challenge in defining a BN is to specify the DAG accurately, and here we focus on algorithms that are used to learn the DAG from observational data since interventional data is less commonly available. There are two main classes of algorithms which can be used to learn graphs from categorical observational data. Constraint-based algorithms, such as the PC algorithm \citep{spirtes1991algorithm}, use statistical tests of marginal and conditional independence to identify dependencies and interdependencies in the data and deduce the possible causal relations from these.

The second major class of learning algorithms consists of score-based algorithms that search over a set of possible graphs and have an objective function which indicates how well each graph visited fits the data. The objective function usually employs either a Bayesian score such as BDeu \citep{heckerman1995learning} which measures how likely the data would be to be generated by that DAG, or an entropy-based score such as BIC \citep{suzuki1999learning} which balances model-fitting with the dimensionality of the model.

In the discrete variable setting considered here, the BDeu score, $S_{BDeu}(G, D)$, for DAG $G$ and dataset $D$, and where the prior probability for all DAGs is the same, is:
\begin{equation} \label{eqn:bdeu}
S_{BDeu}(G, D) = \sum_{i=1}^{n} \sum_{j=1}^{q_i} 
\left[ \log{\frac{\Gamma(\frac{N'}{q_i})}{\Gamma(N_{ij} + \frac{N'}{q_i})}} 
+ \sum_{k=1}^{r_i} \log{\frac{\Gamma(N_{ijk} + \frac{N'}{r_i q_i})} {\Gamma(\frac{N'}{r_i q_i})}} \right]
\end{equation}
where $i$ ranges over $n$ variables, $X_i$,  $j$ ranges over the $q_i$ combinations of values the parents of $X_i$ may take, and $k$ ranges over the $r_i$ possible values of $X_i$. $\Gamma$ denotes the Gamma function, $N_{ij}$ is the number of instances where the parents of $X_i$ have the $j^{th}$ combination of values, and $N_{ijk}$ is the number of instances within those $N_{ij}$ instances where $X_i$ takes the $k^{th}$ value. $N'$ is a hyper-parameter called the \textit{Equivalent Sample Size} (ESS) which expresses the strength of our belief in the uninformative priors for the model parameters which BDeu assumes.

The BIC score, $S_{BIC}(G, D)$, for DAG $G$ and dataset $D$ is:
\begin{equation} \label{eqn:bic}
S_{BIC}(G, D) = \sum_{i=1}^{n} \sum_{j=1}^{q_i} \sum_{k=1}^{r_i}
\left[ N_{ijk} \log{\frac {N_{ijk}} {N_{ij}}} \right]
- K \cdot \frac{\log{N}}{2} \cdot F
\end{equation}
where the symbols take the same values as in equation \ref{eqn:bdeu}, and additionally, $N$ is the total number of instances in the dataset, and $F$ is the total number of free model parameters. The first term in equation \ref{eqn:bic} represents the log-likelihood of the learnt graph $G$ generating the dataset $D$, and represents the model-fitting component of the BIC score. The second term of \ref{eqn:bic} represents the model complexity or dimensionality penalty, with the number of free parameters, $F$, tending to increase as graph complexity does. $K$ is a complexity scaling parameter which can be used to modify the relative weight of the complexity penalty.

The above scores are \textit{decomposable} so that the score for the whole DAG is the sum of the individual scores for each node and its parents, which facilitates computing the scores of neighbouring DAGs. The scores which we use in this study are all \textit{score equivalent} which means that all DAGs in an equivalence class have the same score. This implies that usually some arc orientations cannot be identified when a score equivalent objective function is used.

The score-based algorithms may be further divided into two subgroups. The first subgroup is referred to as \textit{exact} algorithms and guarantees to return the highest scoring graph over a set of graphs, an example being GOBNILP \citep{cussens2011bayesian}. The second subgroup is known as \textit{approximate} algorithms and often involves heuristic search over the search space that returns a local maximum solution. Examples of approximate algorithms include the hill-climbing (HC) algorithm \citep{bouckaert1994properties} and variants of greedy search, as well as genetic algorithms \citep{larranaga1996learning}. The number of possible DAGs rises super-exponentially with the number of variables \citep{robinson1977counting}, and so exact algorithms are typically limited to problems with a few tens of variables. There are also hybrid algorithms which combine constraint-based with score-based solutions. \cite{kitson2023survey} provide a recent review of over seventy algorithms covering score-based, constraint-based and hybrid algorithms.

A learnt CBN is only useful for causal explanation and intervention modelling if it accurately reflects the true causal relationships in the system under study. There are several reasons why the learnt graphical structure might not be an accurate or complete representation of the real world. One group of reasons relates to weaknesses in the algorithms themselves. For example, many approximate score-based algorithms are greedy and end up returning a graph with only a local maximum score. Constraint-based algorithms rely on statistical tests of independence which can only say how unlikely the variables are to be independent, and so mistakes are made in the learning process and these mistakes tend to propagate in the algorithms \citep{spirtes2000causation}. The accuracy of algorithms is also affected by the objective score or independence tests used \citep{scutari2016empirical}, and hyper-parameters which the user must specify such as the independence test threshold p-value, maximum node in-degree, or Equivalent Sample Size (ESS) associated with Bayesian scores \citep{heckerman1995learning}.

A second group of causes of inaccuracy relates to the assumptions that many algorithms rely on, and which frequently do not apply in the real world. These include assuming that there are no missing data values, latent confounders or measurement noise, as well as assumptions about the underlying statistical distributions that the variables belong to. Algorithms often require another assumption known as \textit{faithfulness}, which requires that there are no independencies in the data that are not implied by the DAG. When used in a CBN this is referred to as the \textit{Causal Faithfulness Assumption}

As discussed, correct arc orientation is particularly important for CBNs. As well as the problem of identifying arc orientation in equivalent DAGs, another problem is that different MECs may have rather similar scores, and therefore the confidence in choosing one over the other is low. \cite{koller2009probabilistic} therefore  recommend considering using \textit{Bayesian Model Averaging} (BMA), particularly when learning CBNs. Here, a sampling technique such as Markov Chain Monte Carlo \citep{friedman2003being} or using different classes of learners \citep{constantinou2023open} is used to generate a set of plausible DAGs or CPDAGs. An averaging mechanism can then be used to define the probability of individual features, for instance, arcs.

We focus on networks with discrete categorical variables in this study since these are common in many domain areas such as healthcare, epidemiological data and survey data, for example. There is also a wide range of expert-specified discrete variable networks which provide a basis for making a structural evaluation of the learnt graph which is the focus of this study. We note that there are recent structure learning algorithms which do not support categorical variables and are therefore not considered here. These include additive noise algorithms \citep{peters2014causal} which support continuous variables only and assume a specific functional form for the noise component of the variable values, and continuous optimisation approaches such as NOTEARS \citep{zheng2018dags} which also do not support categorical variables.

\subsection{Related Work}

We define variable ordering to mean the order that the variables appear as columns in the tabular data structure presented to the structure learning algorithm. The variable ordering is arbitrary and so any significant effect it has on learnt graph accuracy is concerning. The fact that variable ordering is important in structure learning has been well-known in the literature for a long time. For example, some early algorithms such as K2 \citep{cooper1992bayesian}, require a topological node ordering to be specified, and there has been considerable interest in determining a good node ordering so that K2 produces accurate graphs. \cite{hruschka2007towards} investigate the use of feature selection as a means of specifying a good ordering, and much more recently \cite{behjati2020improved} construct a cyclic graph where each node has the highest possible scoring set of parents, then recursively examines each cyclic subgraph breaking the cycles to end up with a node order. We also note the recent paper by \cite{reisach2021beware} which shows how the usual method of synthetic data generation for continuous variable networks increases variance down the causal order and thus favours causal discovery by some algorithms over others.

Other approaches which theoretically reduce or eliminate the dependence on variable order include searching node ordering space with approaches such as Order-MCMC \citep{friedman2003being}, genetic algorithms \citep{larranaga1996learning}, and more recently, for example, \cite{bernstein2020ordering} who search ordering space in the presence of latent variables. Some score-based algorithms offer guarantees of returning the highest scoring graph asymptotically as the sample size increases by searching equivalence class space, for example GES \citep{chickering2002optimal}, or the highest scoring graph at a given sample size - for example, the GOBNILP \citep{bartlett2017integer} and A-Star \citep{yuan2011learning} algorithms. Similarly, constraint-based approaches such as PC-Stable \citep{colombo2014order} and the order-based constraint-approach of \cite{bouckaert1992optimizing} are guaranteed to produce the true generating graph providing they are given a complete set of independencies holding in the underlying population.

It might therefore seem that variable ordering is not a pertinent concern, given that many of the above algorithms adopt approaches which mitigate or avoid its effects. However, we note that many of the guarantees given only apply if the required assumptions hold, and that implementation details may affect the algorithm's robustness to variable order. Moreover, practical benchmarks continue to show that the algorithms which we demonstrate are sensitive to variable ordering are competitive when learning structure with realistic sample sizes \citep{scutari2019learns}, or where there is noise in the data \citep{constantinou2021large}. Further, these sensitive algorithms continue to be commonly used in a variety of practical applications. For example, \cite{vitolo2018modeling} have modelled air pollution, \cite{xu2018cognition} and \cite{witteveen2018comparison} cancer, \cite{graafland2020probabilistic} climate, \cite{kitson2021learning} diarrhoea, \cite{graafland2022learning} gene regulatory networks, and \cite{constantinou2023open} COVID-19 using these algorithms. Whilst these papers use a variety of approaches to evaluate the results obtained, including how well the BN fits global probability distribution, the majority do evaluate elements of the graph structure. The structural aspects investigated include the edge structure \citep{graafland2020probabilistic,graafland2022learning}, directed structure \citep{vitolo2018modeling,xu2018cognition,kitson2021learning} and the effects of intervention \citep{constantinou2023open}.

As far as we are aware there has been little work investigating and quantifying the effect of variable ordering on structural accuracy with the greedy-search algorithms that we focus on here. \cite{scutari2019learns} do use five different variable orders when learning graphs from complex real-world climate data to counteract issues of conflicting arc orientations produced in locally dense regions of the graph by constraint algorithms. They do not quantify the effect of variable ordering on structural accuracy, though the results suggest that variable ordering has a smaller effect on the log-likelihood data fitting metric than algorithm choice and the complexity penalty used.

\cite{castelo2003inclusion} describe a variant of hill-climbing where covered arcs are randomly reorientated at each iteration. The intuition is that the algorithm moves between DAGs in the equivalence class at each iteration in a similar way to GES, and is, therefore, less likely to get stuck at a local maximum. The authors report obtaining considerably higher structural accuracy than standard hill-climbing on the Alarm network and some random networks. These covered arcs are the same arcs whose orientation is determined by the variable ordering as we discuss in subsection~\ref{sub:arbitrary} and so their effect on algorithm performance in \citep{castelo2003inclusion} suggests the importance of variable ordering.

Our objectives and contributions are two-fold. Firstly, to understand how variable order affects the learning process in the case of simple hill-climbing, and hence related greedy-search algorithms such as Tabu. Secondly, to quantify the effect of variable ordering and compare it with accuracy variation due to other factors routinely investigated in the literature. We find that the sensitivity to variable ordering often eclipses that due to hyper-parameters, objective function and sample size.

\section{Methodology} \label{sect:methodology}

We evaluate the effect of variable ordering on 16 discrete networks ranging from 8 to 109 variables. Many of these networks are widely used in the literature as case studies to evaluate structure learning algorithms. Table \ref{tab:networks} lists them and their key characteristics. The Sports, Property, Formed and Diarrhoea networks are available from the Bayesys repository \citep{bayesysrepository}, and the rest are obtained from the bnlearn repository \citep{bnrepository}. The network definitions specify the causal structure and the Conditional Probability Tables (CPTs)\footnote{Some CPTs are modified for six variables in Water, two variables in Barley, two variables in Win95pts and six variables in Pathfinder to reduce the risk of single-valued variables at low sample sizes. The probability modifications across these variables are never larger than 0.003. The purpose of this is to reduce the risk of generating single-state variables at lower sample sizes. Single-state variables are where the variable takes the same value for all rows in the dataset. The algorithms studied here do not support such datasets.} for the discrete variables. We use these definitions to generate synthetic datasets with sample sizes of \{10, 20, 40, 50, 80, 100, 200, …., 8x10\textsuperscript{6}, 10\textsuperscript{7}\}. The default variable ordering of these datasets is alphabetic. Table~\ref{tab:networks} presents the number of variables and arcs in each network, as well as mean and maximum degree and in-degree. The final column shows what fraction of the arcs are reversible, that is, those edges which are undirected in the CPDAG.

\begin{table}[ht]
\centering
\begin{tabular}{lcccccccc}
\hline
\thead{Network} & \thead{Number \\ of \\ variables} & \thead{Number \\ of \\arcs} & \thead{Mean \\ in-degree} & 
\thead{Maximum \\ in-degree} & \thead{Mean \\ degree} & \thead{Maximum \\ degree} & 
\thead{Fraction of \\ arcs which \\ are reversible} \\
\hline
asia & 8 & 8 & 1 & 2 & 2 & 4 & 0.375\\
sports & 9 & 15 & 1.67 & 2 & 3.33 & 7 & 0.867\\
sachs & 11 & 17 & 1.55 & 3 & 3.09 & 7 & 1\\
child & 20 & 25 & 1.25 & 2 & 2.5 & 8 & 0.48\\
insurance & 27 & 52 & 1.93 & 3 & 3.85 & 9 & 0.346\\
property & 27 & 31 & 1.15 & 3 & 2.3 & 6 & 0.097\\
diarrhoea & 28 & 68 & 2.43 & 8 & 4.86 & 17 & 0.412\\
water & 32 & 66 & 2.06 & 5 & 4.12 & 8 & 0.091\\
mildew & 35 & 46 & 1.31 & 3 & 2.63 & 5 & 0\\
alarm & 37 & 46 & 1.24 & 4 & 2.49 & 6 & 0.087\\
barley & 48 & 84 & 1.75 & 4 & 3.5 & 8 & 0.107\\
hailfinder & 56 & 66 & 1.18 & 4 & 2.36 & 17 & 0.258\\
hepar2 & 70 & 123 & 1.76 & 6 & 3.51 & 19 & 0.073\\
win95pts & 76 & 112 & 1.47 & 7 & 2.95 & 10 & 0.107\\
formed & 88 & 138 & 1.57 & 6 & 3.14 & 11 & 0.174\\
pathfinder & 109 & 195 & 1.79 & 5 & 3.58 & 106 & 0.626\\
\hline
\end{tabular}
\caption{Networks used in this study}\label{tab:networks}
\end{table}

We investigate the effect of variable ordering on two score-based algorithms. The first of these, HC, starts from an empty DAG and compares the scores of each neighbouring DAG, and moves to the neighbouring DAG with the highest score at each iteration. The process continues until there is no neighbouring DAG which improves on the current DAG’s score. In the simple variant described here, neighbouring DAGs are defined as ones created by making a single arc addition, deletion or reversal to the current DAG. The final DAG is returned as the learnt DAG, and it generally has just a local maximum score. Thus HC is an approximate algorithm. Variants of the basic HC algorithm have been proposed to try to escape local minima, notably restarts with randomised initial DAGs, and the Tabu list which allows for a limited number of changes that minimally decrease the objective score \citep{bouckaert1994properties}. The Tabu approach is the second score-based algorithm investigated here. 

Three constraint-based algorithms are also investigated. The first of these is the PC-Stable algorithm \citep{colombo2014order}, which starts with a complete undirected graph and performs marginal and Conditional Independence (CI) tests to remove edges from the graph. For efficiency, the algorithm starts with marginal independence tests and moves onto conditional independence tests with conditioning sets of increasing size. This results in an undirected graph, called the \textit{skeleton}, where each edge represents a direct dependency between variables. Conditional dependence tests are then used to orientate as many edges in the skeleton as possible. Earlier versions of PC were sensitive to variable ordering, but PC-Stable claimed to eliminate this problem. The PC-Stable algorithm considers independencies between all variables and constructs the graph “globally”, whereas local constraint-based algorithms construct the structure relating to each node separately first, and then merge these to construct the final graph. The local constraint-based GS \citep{margaritis1999bayesian} and Inter-IAMB \citep{tsamardinos2003time} algorithms are also included in this study. 

Two hybrid algorithms are investigated as well. These have an initial local constraint-based algorithm which determines the skeleton of the graph, and then a second HC phase which only considers adding arcs consistent with that skeleton. The first hybrid algorithm MMHC \citep{tsamardinos2006max} uses a local constraint-based algorithm MMPC \citep{tsamardinos2003time} in its first constraint-based phase, and the second one, H2PC, uses HPC (Hybrid Parents and Children) \citep{gasse2014hybrid} in that phase.

The results in this study are obtained using the algorithm implementations provided in version 4.7 of the bnlearn package \citep{scutari2009learning,bnlearn}. We use the default objective scores, conditional independence test functions and hyper-parameters\footnote{For score-based and hybrid algorithms, the default is to use the BIC score with a complexity scaling of 1. For constraint-based algorithms and hybrid algorithms, the Mutual Information Conditional Independence test is used with a p-value threshold of 0.05.} except for experiments where we explicitly report changing one of those factors. The constraint-based algorithms investigated generally produce a CPDAG\footnote{In a small minority of cases the constraint-based implementations we use to return a PDAG which cannot be completed into a CPDAG, an issue noted by} \cite{scutari2019learns}.  In these cases, we compare the learnt PDAG with the reference CPDAG, whereas the score-based and hybrid algorithms produce a DAG. To get consistent evaluations across all algorithm classes, for score-based and hybrid algorithms we construct the CPDAG to which the DAG belongs, and compare this with the CPDAG to which the reference DAG belongs.

The learnt and reference CPDAGs are primarily compared using the F1 metric. The F1 score is widely used in the literature and is intuitive in that it ranges from 0 to 1 where a higher score represents closer agreement between the learnt and reference graph. The F1 score is comparable across networks with different numbers of variables. Moreover, our prime motivation for this study is to investigate the effect of variable ordering on graph structure, and hence causal discovery and intervention modelling. Thus evaluation using a structural measure such as F1 is appropriate. The F1 metric and CPDAG extension methods provided in bnlearn \citep{scutari2009learning,bnlearn} are used. The F1 used is the simple harmonic mean of Precision and Recall:
\begin{equation}
F1 = \frac{2 \times Precision \times Recall}{Precision + Recall}
\end{equation}
where Precision and Recall are defined as
\begin{equation}
Precision = \frac{TP}{TP + FP}, \, \, Recall = \frac{TP}{TP + FN}
\end{equation}
and TP is the number of True Positives, FP is False Positives and FN is False Negatives. Table~\ref{tab:metrics} shows the contribution that the possible combinations of edge types in the learnt and true graphs make to the True Positive, False Positive and False Negative counts used to compute precision and recall, and hence the F1 metric used in this study. Results using some additional metrics, the Structural Hamming Distance (SHD) \citep{tsamardinos2006max}, and edge characterisation are provided in Appendix \ref{app:supplementary} and their derivation is also shown in Table~\ref{tab:metrics}.

\begin{table}
\centering
\begin{tabular}{ccccccc}
\hline
\thead{Learnt \\ graph} & \thead{True \\ graph} & \thead{TP} 
& \thead{FP} & \thead{FN} & \thead{SHD} & \thead{Learnt edge \\ characterisation}\\
\hline
$\longrightarrow$ & $\longrightarrow$ & 1 & 0 & 0 & 0 & same \\
\textbf{\large{\textemdash}} & \textbf{\large{\textemdash}} & 1 & 0 & 0 & 0 & same \\
$\longrightarrow$ & no edge & 0 & 1 & 0  & 1 & extra \\
\textbf{\large{\textemdash}} & no edge & 0 & 1 & 0 & 1 & extra \\
no edge & $\longrightarrow$ & 0 & 0 & 1 & 1 & missing \\
no edge & \textbf{\large{\textemdash}} & 0 & 0 & 1 & 1 & missing \\
$\longrightarrow$ & $\longleftarrow$ & 0 & 1 & 1 & 1 & misorientated \\
$\longrightarrow$ & \textbf{\large{\textemdash}} & 0 & 1 & 1 & 1 & misorientated \\
\textbf{\large{\textemdash}} & $\longrightarrow$ &  0 & 1 & 1 & 1 & misorientated \\
\hline
\end{tabular}
\caption{The contribution to the True Positive, False Positive, False Negative counts (and hence F1) and SHD resulting from different combinations of edges in the learnt and true graph. The characterisation of incorrect edges in the learnt graph is also shown. The formulations in this table follow those in the widely-adopted bnlearn software \citep{bnlearn}.}
\label{tab:metrics}
\end{table}

\section{Results}

\subsection{Arbitrary DAG Changes in the HC Algorithm}
\label{sub:arbitrary}

\begin{figure}
     \centering
     \begin{subfigure}[b]{0.415\textwidth}
         \centering
         \includegraphics[width=\textwidth]{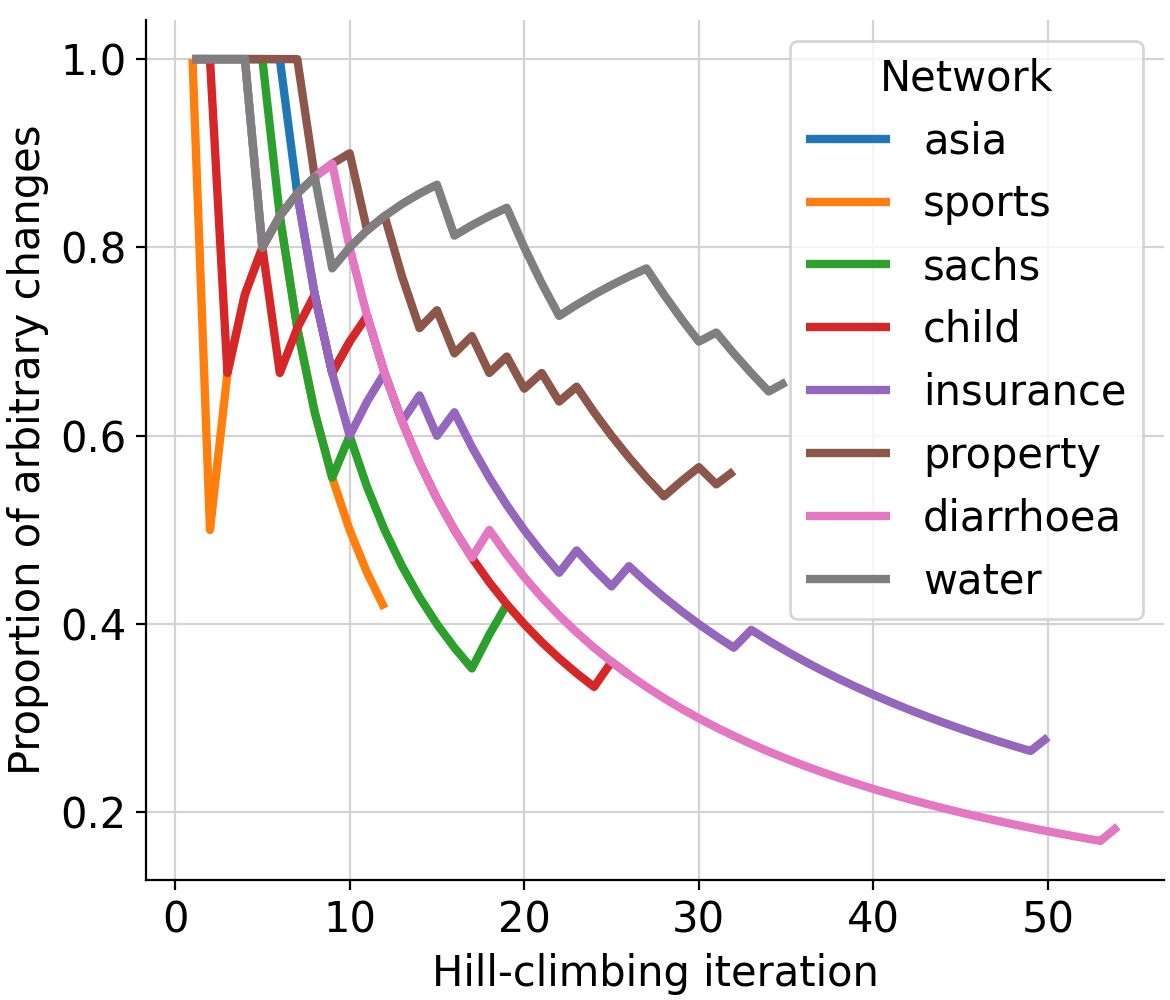}
         \caption{Smaller networks}
         \label{fig:hc_arbitrary_sm}
     \end{subfigure}
     \hfill
     \begin{subfigure}[b]{0.565\textwidth}
         \centering
         \includegraphics[width=\textwidth]{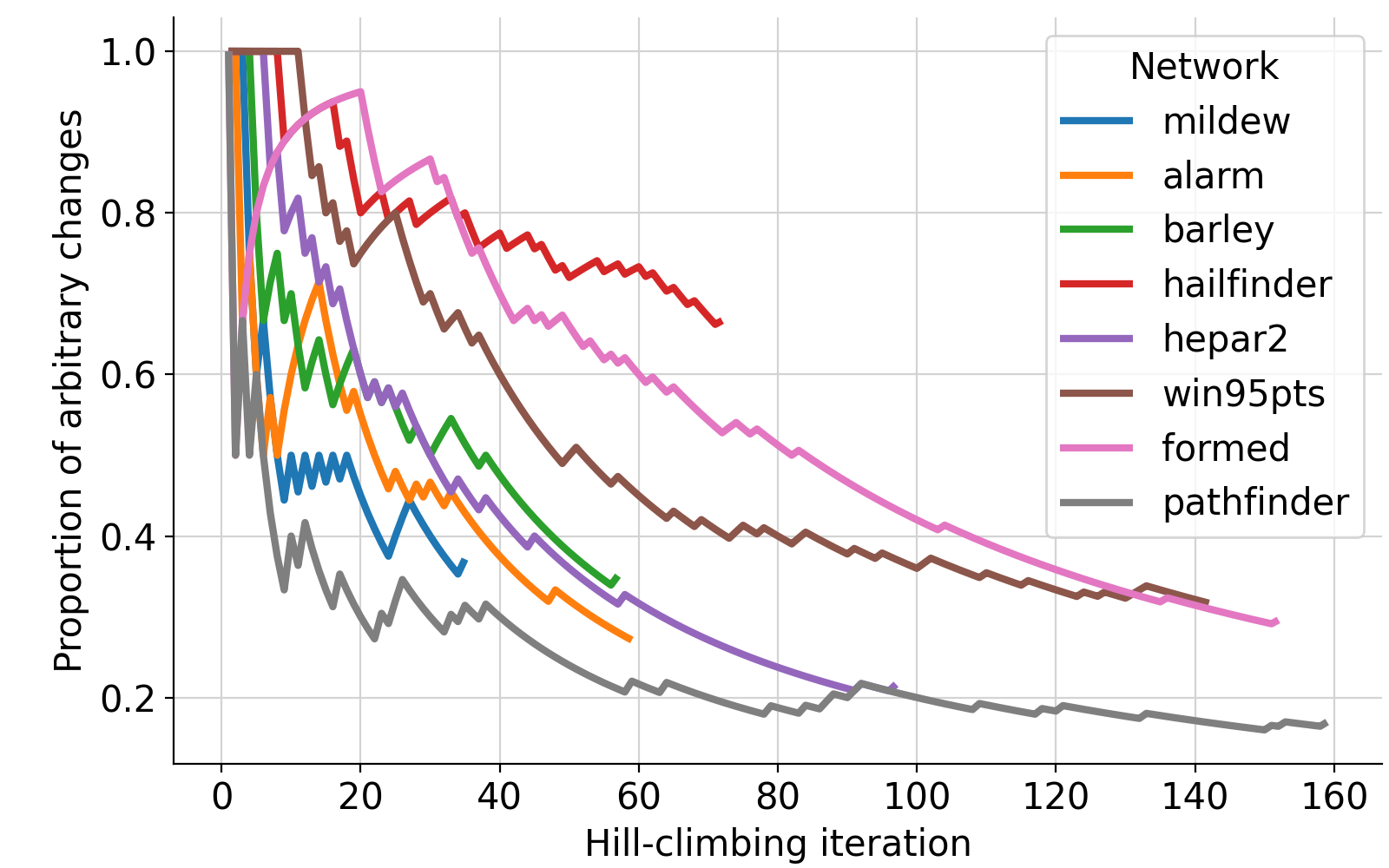}
         \caption{Larger networks}
         \label{fig:hc_arbitrary_lg}
     \end{subfigure}
        \caption{Arbitrary DAG changes in HC algorithm using 10,000 rows of data}
        \label{fig:hc_arbitrary}
\end{figure}

We first examine the individual changes - arc addition, reversal or removal - which the HC algorithm makes at each iteration as it learns the DAG structure. In particular, we note where changes are arbitrary; that is, where two neighbouring DAGs are Markov equivalent. Figure~\ref{fig:hc_arbitrary} shows the proportion of graphical modifications which are arbitrary for each network with a sample size of 10,000. The first arc added must always be in an arbitrary direction, and hence all lines on the chart start with a proportion of 1.0 arbitrary changes at iteration 1. For the Pathfinder dataset in Figure~\ref{fig:hc_arbitrary_lg}, the greatest score improvement at iteration 2 happens to be provided by adding an arc onto the first arc to create a chain. The alternative orientation of that edge, which would create a collider, has a smaller score improvement so this change is not arbitrary. Thus, the proportion of arbitrary changes at iteration 2 for Pathfinder drops to 0.5. The next highest scoring change is an arc addition that does not connect onto this initial chain. Hence that change is arbitrary too and so the proportion of arbitrary changes rises to 2/3 at iteration 3.  

We can contrast this behaviour with that of Hailfinder in Figure~\ref{fig:hc_arbitrary_lg}. In that case, the first eight highest-scoring arc additions are all between completely separate pairs of nodes, so that the DAG at iteration eight consists of eight unconnected arcs. Thus, at that iteration, the proportion of arbitrary arcs remains at 1.0. The highest scoring change at iteration 9 happens to be an arc addition which creates a collider with one of the existing arcs and so this non-arbitrary change reduces the proportion to 8/9. The following seven changes are also arbitrary. 

Figure~\ref{fig:hc_arbitrary} thus provides an overview of the HC learning process and the proportion of edge modifications whose orientation is determined arbitrarily by the variable order. Based on these results, it is reasonable to conclude that variable ordering has a significant effect in the initial iterations and that part of this effect might propagate to the final graph.

\subsection{The Impact of Variable Ordering on HC Algorithm Graphs} \label{sub:hc_ordering}

\begin{figure}[htp]
    \centering
    \includegraphics[width=14cm]{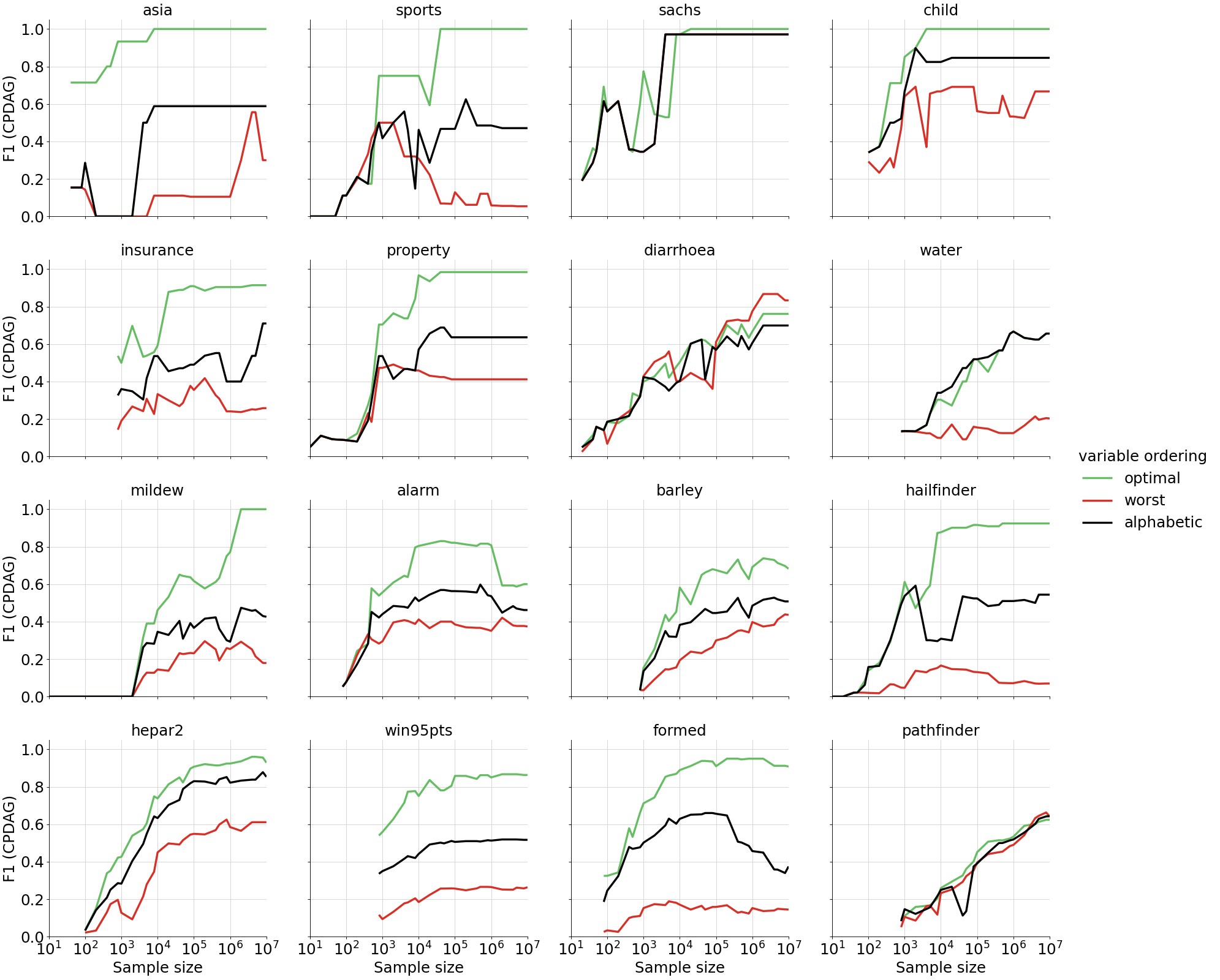}
    \caption{F1 against sample size for each variable ordering and network for the HC algorithm. Each plot starts at the sample size at which there are no single-valued variables. (Note that the red and black lines are coincident for the Sachs network which is why the former is not visible).}
    \label{fig:hc_ordering}
\end{figure}

We compare the sensitivity of the F1 score relative to the default alphabetic ordering and two other orderings which we term “optimal” and “worst”. In optimal ordering, the variables are ordered so that they are consistent with the topological ordering of the nodes in the reference graph. This optimal ordering ensures that the algorithm always chooses the correct orientation when faced with adding an arc where the two orientations give the same score improvement. Worst ordering is the reverse of optimal and ensures that the algorithm always makes the wrong decision when adding an arc in the same circumstances.

Figure~\ref{fig:hc_ordering} illustrates how the F1 score for each variable ordering changes, as we vary the sample sizes for each of the 16 networks. For many networks, we see that variable ordering makes a large difference in learnt accuracy. In the most extreme cases, such as Asia, Formed, Property, and Hailfinder, there is a difference of more than 0.5 in F1 at large sample sizes. These networks tend to be those which have a high proportion of arbitrary changes as shown in Figure~\ref{fig:hc_arbitrary}. In contrast, there are some networks, particularly Pathfinder, Diarrhoea and Hepar2, where the variable ordering has much less influence on accuracy. These tend to be networks with a smaller proportion of arbitrary changes as the DAG is being learnt. The performance of the alphabetic ordering can be viewed as a random ordering, which is why the scores produced for this ordering are mostly between that of the worst and optimal ordering.

Figure~\ref{fig:hc_ordering} also provides a comprehensive view into how HC’s accuracy varies over a wide range of sample sizes and networks. The broad trend for most networks is that accuracy rises with sample size but then tends to reach a plateau. The height of this plateau varies with the variable ordering and with the network, as does the sample size at which accuracy begins to level off. For some smaller networks such as Asia, Child and Property, maximum accuracy is achieved when sample sizes are of the order of 10\textsuperscript{3} to 10\textsuperscript{4} rows, whereas larger networks such as Win95pts obtain maximum accuracy at higher sample sizes. Interestingly, some networks such as Water and Pathfinder have not yet plateaued at 10\textsuperscript{7} rows. The picture is also complicated because there are some cases where accuracy drops as sample size increases, for example, the Formed and Alarm networks using alphabetic ordering. The variation of F1 with sample size is likely related to network complexity and strength of the implied relationships, though the relationship is probably not simple. For instance, we note that an F1 of 1.0 is achieved at large sample sizes with optimal ordering for the Mildew network which is relatively complex having 540,150 free parameters and 2.63 edges per variable. Whereas, the less complex Alarm network with 509 free parameters and 2.49 edges per variable achieves only an F1 of around 0.6 at the largest sample sizes.

Another issue is the erratic behaviour of some of the plots where there are some large increases or, less often, decreases which occur between two sample sizes. Examples include the Sachs network at around 2,000 rows and Mildew at 10\textsuperscript{6} rows. It is suggested that possibly some differences in the learning process have a ‘cascading’ effect which causes the sequence of changes to be dramatically altered. Further research in this area would be useful.

Figures \ref{fig:ord_hc_opt_edges} and \ref{fig:ord_hc_worst_edges} in Appendix~\ref{app:supplementary} characterise incorrect edges in the learnt graph when optimal and worst variable ordering is used respectively. These figures are based on the same experiments used for Figure~\ref{fig:hc_ordering}, and use the edge characterisations defined in Table~\ref{tab:metrics}. The number of edges is scaled by the number of edges in the true graph, and the F1 value is also shown. Broadly speaking, the trend of missing edges is the same for both orderings over all networks; that is, it drops as sample size increases. This may be because weaker dependency relationships overcome the BIC complexity penalty as the sample size increases. Perhaps unsurprisingly, there are generally many more 'misorientated' edges when worst ordering is used compared to optimal ordering, and this number tends to increase with sample size as more edges are discovered. Interestingly, where there are large numbers of misorientated edges, for example, for the hailfinder, win95pts and formed networks in Figure~\ref{fig:ord_hc_worst_edges}, there are considerably more extra edges. This suggests variable ordering can also affect the skeleton of the learnt graph.

Figure \ref{fig:ord_hc_shd} plots the SHD value for the same set of experiments to facilitate comparison with studies which report the SHD values. Again, as expected, this mirrors the structural differences shown in Figure~\ref{fig:hc_ordering}, with the green plot for the optimal ordering usually being the lower line as the more structurally accurate the learnt CPDAG is, the lower the SHD value.

Whilst the focus of this study is evaluating structural accuracy, Figure~\ref{fig:ord_hc_loglik} shows the log-likelihood of the CPDAG learnt with each ordering. The log-likelihood scales linearly with the sample size, so this figure shows the log-likelihood divided by sample size so that it is comparable at different sample sizes. There is much less variation between the orderings than is seen in the structural measures such as F1. Nonetheless, there is some notable variation for some networks, for example, child, insurance and win95pts.

\begin{figure}[htp]
    \centering
    \includegraphics[width=14cm]{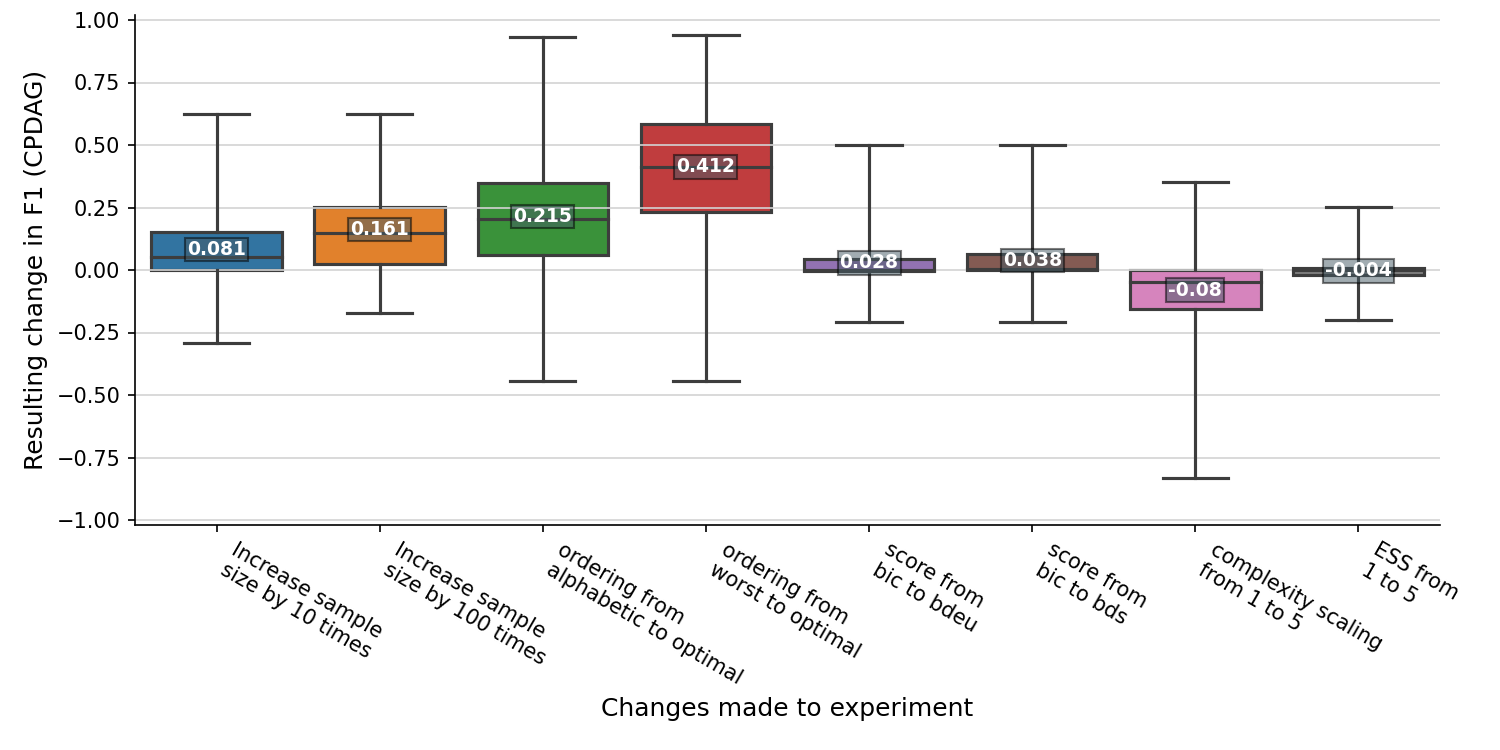}
    \caption{Impact on F1 scores (CPDAG) of changing sample size, variable ordering, score or hyper-parameters across all networks using the HC algorithm with sample sizes between $10^3$ and $10^6$. Each plot shows the mean change as a number, the median change as a horizontal black line, the interquartile range as the coloured rectangle, and the minimum and maximum values as whiskers.}
    \label{fig:hc_impact}
\end{figure}

Figure~\ref{fig:hc_impact} summarises the impact of variable ordering on the accuracy of HC and compares this with the impact of other factors which affect learning accuracy. Note that for this figure and all subsequent results, we only consider sample sizes in the range 10\textsuperscript{3} to 10\textsuperscript{6}. This allows us to use the same range of sample sizes across all networks and algorithms\footnote{This is because, within this range, all networks have no single-valued variables and all algorithms considered can learn graphs within a limit of 24 hours elapsed time.}

Figure~\ref{fig:hc_impact} presents a series of boxplots showing the distribution of changes to F1 accuracy, where one factor is changed at a time. For example, the leftmost, dark blue plot shows the change in accuracy when the sample size is increased tenfold. To compute this, the F1 from sample size 1,000 is compared with that of sample size 10,000, sample size 2,000 compared with 20,000 and so on, over sample sizes and networks. Similarly, we compare accuracy from sample sizes which are a multiple of 100 times the other in the orange boxplot.

The remaining plots show the impact where the sample size is kept the same but where another factor is varied. The green plot shows the change in accuracy resulting from using the optimal ordering rather than the alphabetic one, and the red plot shows the change in accuracy moving from the worst to optimal ordering. The last set of plots in Figure 3 shows the impact of some hyper-parameter or objective score changes. These examine the effect of changing the objective score from BIC to BDeu and changing the objective score from BIC to a more recent Bayesian score, BDS \citep{scutari2016empirical}. The last two plots show the impact of changing the BIC complexity scaling factor, $K$ in equation \ref{eqn:bic}, from 1 to 5 and the BDeu Equivalent Sample Size (ESS), $N'$ in equation \ref{eqn:bdeu}, from 1 to 5. These changes in the complexity scaling and ESS have a considerable impact on the mean number of edges per variable, or \textit{graph density}, changing it by 20\% and 17\% respectively. The magnitude of the changes to these hyper-parameters reflects the scale of changes that might be made in real-world practice.

Figure~\ref{fig:hc_impact} illustrates that changing the variable ordering from worst to optimal has the largest overall impact on F1, with a mean improvement of 0.412 and an interquartile range from 0.233 to 0.584. The mean improvement going from alphabetic to optimal ordering is 0.215, with an interquartile range from 0.062 to 0.348. Comparing this with improvements due to other factors, Figure~\ref{fig:hc_impact} illustrates that changing from alphabetic to optimal ordering results in a higher mean improvement than increasing the sample size by one hundred times (mean improvement 0.161), and considerably higher improvement than increasing sample size by ten times. This observation includes cases where increasing sample size worsens F1 corresponding to the negative gradients seen in Figure~\ref{fig:hc_ordering}. Changes to F1 due to changing the objective score from BIC to BDeu and from BIC to BDS have small mean changes of 0.028 and 0.038 respectively. Changing the BIC complexity penalty multiplier hyper-parameter and BDeu ESS from 1 to 5 produces a mean change in F1 of -0.080 and -0.004 respectively. Thus, the mean changes and interquartile ranges associated with changing the objective score or hyper-parameters are smaller than the changes associated with changing the sample size and considerably smaller than the changes associated with changing from optimal to worst or alphabetical variable ordering.

The comparative impact of variable ordering in low-dimensional settings is shown in Figure~\ref{fig:ord_hc_impact_lowd} in Appendix~\ref{app:supplementary}. This figure is constructed in the same way as Figure~\ref{fig:hc_impact} but is based on sample sizes between $10$ and $10^3$. The impact of increasing sample size is now larger, with the impact of increasing sample size by 100 times now 0.360 compared to 0.161 for sample sizes $10^3$ to $10^6$. In contrast, the impact of changing from the worst to optimal ordering has fallen from 0.394 to 0.166 in the low-dimensional setting. Thus, sample size has more impact than variable ordering in low-dimensional settings, though the effect of variable ordering remains considerable, and changing from worst to optimal ordering continues to have more impact than the hyper-parameter changes.

\subsection{The Effect of Variable Ordering on other Structure Learning Algorithms}

\begin{figure}[htp]
    \centering
    \includegraphics[width=12cm]{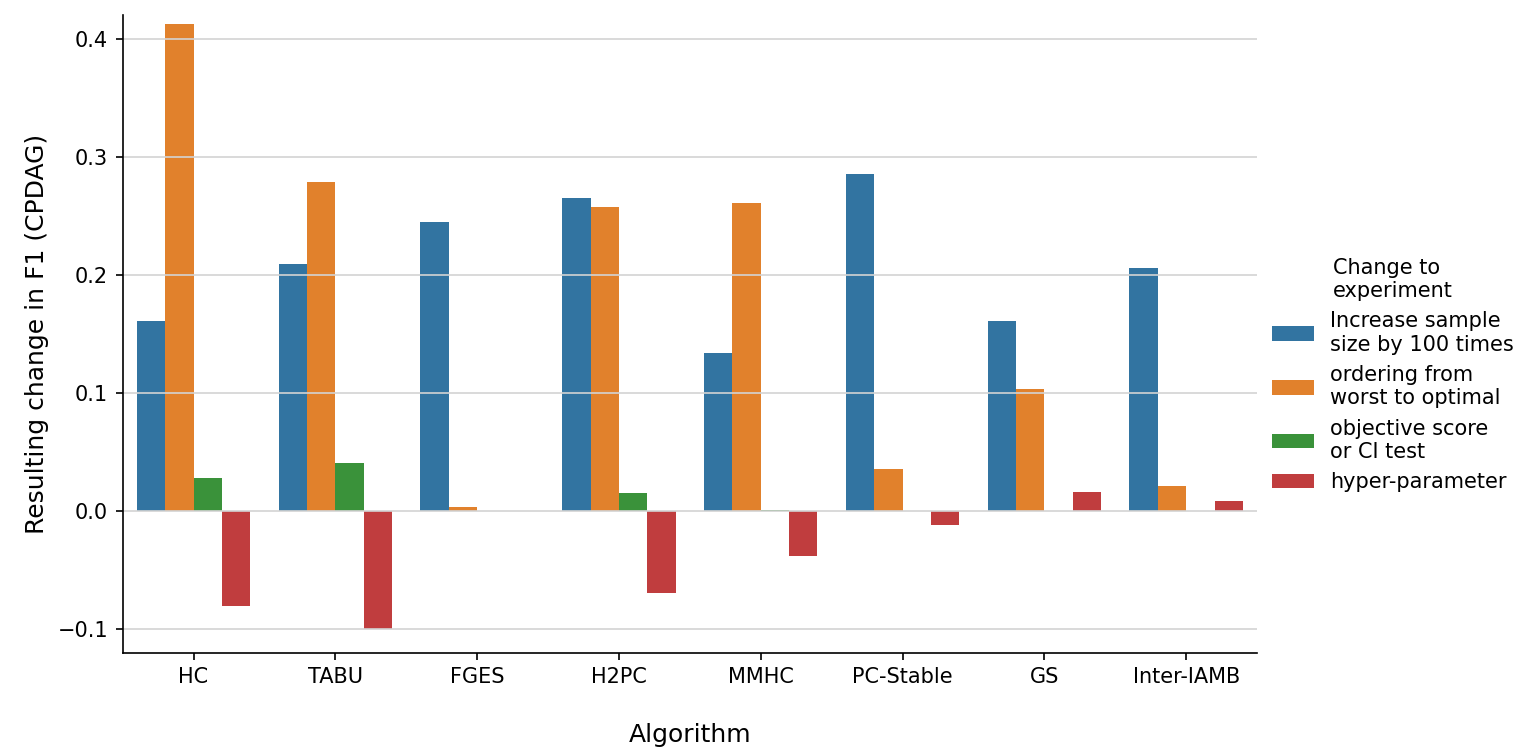}
    \caption{The sensitivity of different algorithms to variable ordering compared to other factors in terms of impact on F1 score (CPDAG). The comparisons are: sample size increased by 100 times; variable ordering changed from worst to optimal; score function changed from BIC to BDeu for score-based and hybrid algorithms or the CI test from Mutual Information to Chi-squared for constraint-based algorithms; and, the complexity scaling hyper-parameter is changed from 1 to 5 for score-based and hybrid algorithms or the p-value significance hyper-parameter changed from 0.05 to 0.01 for constraint-based algorithms.}
    \label{fig:algorithms}
\end{figure}

Figure~\ref{fig:algorithms} shows the sensitivity to variable ordering for the algorithms described in section~\ref{sect:methodology} and compares it with their sensitivity to other selected factors (details are given in the figure caption). TABU is a variant of HC and is, as one might expect, sensitive to variable ordering, but the mean F1 change of 0.278 is considerably less than that of 0.412 for HC. This may be because Tabu undoes some of the incorrect arbitrarily orientated arcs as it escapes local maxima towards the end of the learning process. One would expect hybrid algorithms such as H2PC and MMHC, which utilise the HC algorithm in their score-based learning phase to be affected by variable ordering also, and indeed this is the case. However, the constraint-based phase appears to moderate the effect of variable ordering, so that H2PC and MMHC have mean F1 changes of 0.256 and 0.261 respectively.

Overall, variable ordering is found to be the most impactful factor on F1 score in three out of the seven algorithms investigated, and the second most impactful factor for the remaining four algorithms. For score-based and hybrid algorithms, variable order has a larger impact than whether BIC or BDeu score is used or setting the BIC complexity penalty to 5.

Figure~\ref{fig:algorithms} also shows that the constraint-based algorithms are sensitive to variable ordering, with a mean change in F1 of 0.036, 0.104 and 0.021 for PC-Stable, GS and Inter-IAMB respectively. Although these sensitivities are small relative to those observed in other algorithms, they are still larger than the sensitivity of the constraint-based algorithms to the change in the CI test and p-value threshold investigated. Note that because constraint-based learning tends to involve dependency tests across sets of variables in triples, it might be possible that other variable orderings have a larger effect on constraint-based algorithms. The sensitivity of PC-Stable is surprising given that it is designed not to be sensitive to variable ordering. On the other hand, we note that we find large sensitivities in algorithms outside the PC family, which does not seem to have been considered much in the literature. 

\begin{table}
\centering
\begin{tabular}{lcccc}
\hline
\thead{Algorithm} & \thead{Worst ordering} & \thead{Alphabetic ordering} & \thead{Optimal ordering} & \thead{Random ordering}\\
\hline
TABU & 0.130 & 0.104 & {\color{red} -0.004} & 0.080 \\
H2PC & 0.048 & {\color{red} -0.009} & {\color{red} -0.107} & {\color{red} -0.001} \\
MMHC & {\color{red} -0.066} & {\color{red} -0.118} & {\color{red} -0.218} & {\color{red} -0.106} \\
PC-Stable & 0.158 & {\color{red}-0.032} & {\color{red}-0.222} & {\color{red}-0.008} \\
GS & {\color{red}-0.074} & {\color{red}-0.252} & {\color{red}-0.383} & {\color{red}-0.243} \\
Inter-IAMB & 0.111 & {\color{red}-0.100} & {\color{red}-0.280} & {\color{red}-0.066} \\
\hline
\end{tabular}
\caption{Mean F1 scores (CPDAG) of the other algorithms relative to HC, across all three variable orderings and, in the final column, across a set of ten random orderings. A positive discrepancy indicates superior performance relative to HC, and vice-versa.}
\label{tab:rankings}
\end{table}

Table~\ref{tab:rankings} shows the mean F1 scores that each of the other algorithms generate relative to the F1 scores of HC, averaged across all networks and sample sizes, for each of the three variable orderings. In addition, the final column in the table presents results where ten random variable orders are generated for each network, and results averaged over these ten random orderings for each algorithm, network and sample size. Table~\ref{tab:rankings} thus provides a ranking of the algorithms for different variable orderings. We see that HC is the highest-ranked algorithm when using an optimal ordering since the other six algorithms produce lower F1 scores than HC for that ordering, but the rankings change for the other two variable orderings. The alphabetic and random ordering show similar results, with Tabu more accurate than HC, H2PC and PC-Stable broadly on a par with HC, but MMHC, GS and Inter-IAMB less accurate to varying degrees. Overall, variable ordering is found to also affect algorithm ranking considerably.

\section{Conclusions}

This study examines the impact that the arbitrary variable ordering within the dataset has on the accuracy of graphs learnt by commonly used structure learning algorithms using discrete categorical data. Whilst the importance of some aspects of variable ordering is well known, we are unaware of any other study that quantifies its impact on discrete variable networks.

We show that the impact on some commonly used and competitive approximate score-based algorithms which search in DAG-space is considerable, and noteworthy effects are also found in some hybrid and constraint-based algorithms. We recognise that this sensitivity is unlikely to arise in score-based algorithms which search ordering space or are exact for example, but it \textit{does} affect some well-established and state-of-the-art algorithms which are commonly used in practical applications \citep{vitolo2018modeling, witteveen2018comparison, graafland2020probabilistic, sattari2021application} and found to be the best performers in recent comparative evaluations \citep{scutari2019learns,constantinou2021large}.

By examining the way a DAG develops iteration by iteration in the simple HC algorithm, we find that arbitrary decisions about edge modifications play an important role in determining the accuracy of the learnt graph and thus, in judging the structure learning capability of an algorithm. This is particularly so when HC and its variants start from an empty graph and when many of the first arcs added are initially not connected to each other. This means a higher proportion of the initial arc orientations are arbitrary. The magnitude of this effect depends upon the underlying causal graph so that some networks are much more sensitive to variable ordering than others.

We start by investigating the bnlearn implementation of HC which is widely used in the literature \citep{scutari2009learning,bnlearn} and find that these arbitrary edge orientations are made on the basis of arbitrary variable ordering in the dataset, which therefore has an impact on the accuracy of the learnt CPDAG. For HC, variable ordering is found to typically have a larger effect than sample size, objective score or hyper-parameters used. We also examine the effect of variable ordering on score-based TABU, the more recent hybrid MMHC and H2PC which use HC in their score-based phase, as well as on constraint-based PC-Stable, Inter-IAMB, and GS. We find that whilst variable ordering has a smaller effect on these algorithms compared to the effect it has on HC, it is still considerable and often stronger than the effect of other factors such as sample size, objective score, and hyper-parameter changes.

Whilst it might be argued that these results are somewhat specific to this particular bnlearn implementation of score-based or constraint-based learning, simple hill-climbing search has to make these choices on some arbitrary basis and so these concerns are likely to be relevant to other implementations, as well as in other algorithms. We find smaller sensitivity for the constraint-based algorithms, though still some, which is surprising in the case of PC-Stable which is designed not to be sensitive to variable ordering.

This study investigates structure learning from discrete data. However, these findings are likely to be relevant to the algorithms we studied when learning from continuous data. If a score-equivalent objective function is used, then this too would involve making arbitrary arc orientations of the kind we illustrate here. Extending this work to cover continuous data could be a valuable exercise, especially because many of these algorithms have been shown to recover considerably denser graphical structures when trained with continuous data \citep{constantinou2023open}, compared to discretising the data, and it remains unclear how this might affect the sensitivity to the ordering of the variables. We note too that Bayesian Model Averaging is recommended for causal discovery and applying this to graphs learnt from different variable orders to see if it mitigates their influence would be interesting.

This work has focused on the effect of variable ordering on DAG structure but supplementary results with simple-hill-climbing demonstrate that the log-likelihood score of the learnt graph and skeleton are affected too, suggesting that probability distribution inference would also be impacted. Quantifying this effect, particularly when predicting the marginal distribution of key variables, and the extent to which it affects other algorithms could be another fruitful direction for further research.

The effect of variable ordering on structure learning might be well-known for some in the research community, but we argue that its importance is largely underestimated. This is because whilst it is typical for structure learning algorithms to be assessed across different objective functions, varied sample sizes, and different hyper-parameters to convince readers about their validity, almost none of the algorithms published in the literature is tested for sensitivity to variable order. This paper suggests that variable ordering has a much stronger effect on structure learning performance than previously assumed, especially when compared to the other factors on which algorithms are tested for sensitivity. This in turn raises questions about the validity of algorithms which are sensitive to variable ordering, and also comparative benchmarks which do not take the effect into account.

\backmatter

\section*{Declarations}

\textbf{Conflict of Interest:} The authors have no competing interests to declare that are relevant to the content of this article.

\bibliography{kitson24a}

\appendix

\section{Supplementary Results}
\label{app:supplementary}

This appendix contains supplementary results which present metrics other than the F1 metric which is the focus of this paper, or which examine a subset of the experiments to illustrate a specific point made in the main text.

\begin{figure}[htp]
    \centering
    \includegraphics[width=13cm]{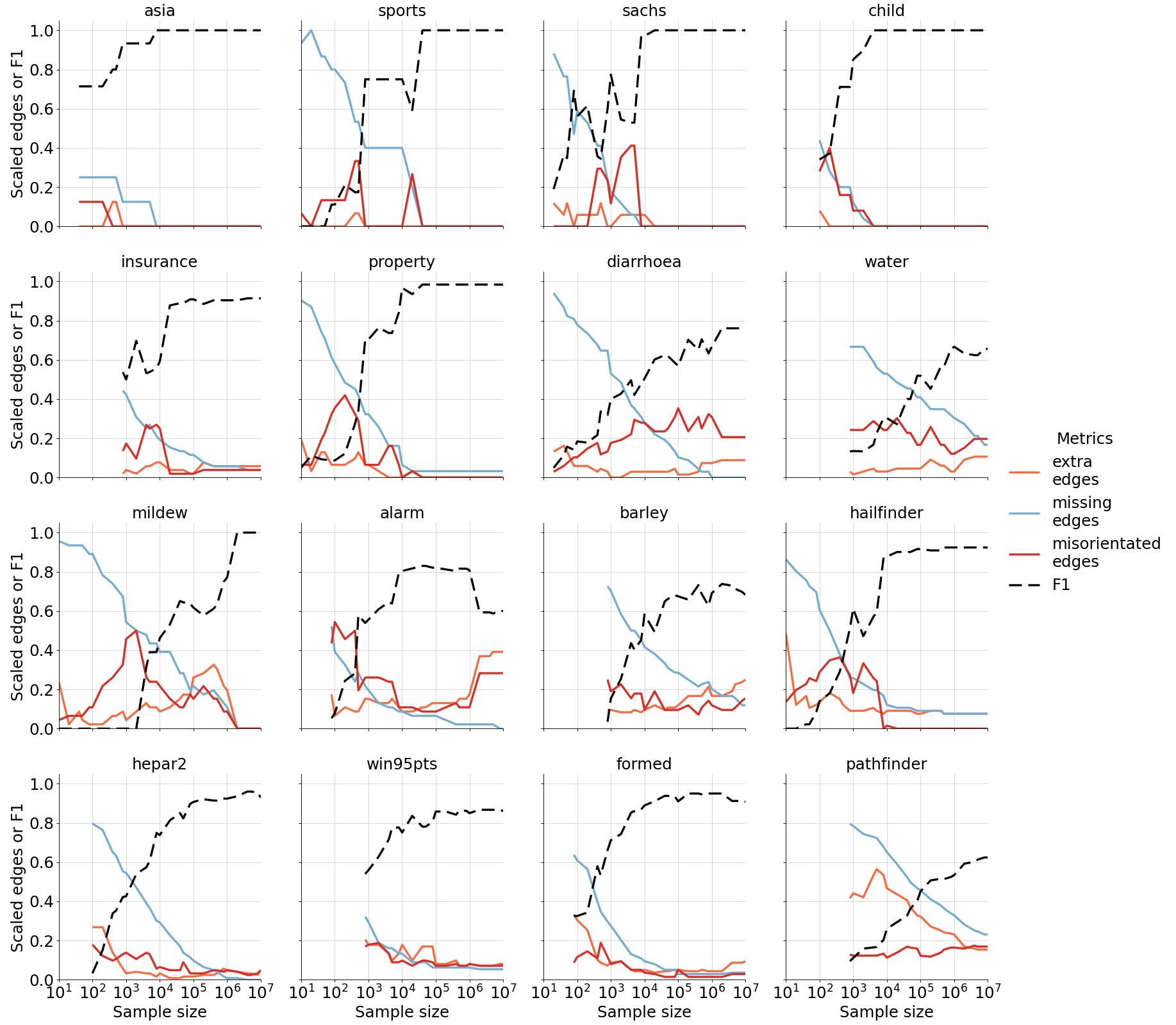}
    \caption{Scaled number of incorrect edges and F1 of learnt CPDAG when \textbf{optimal ordering} is used. The red line shows extra edges, either directed or undirected, the blue line shows missing edges and the orange line shows edges which are misorientated. Misorientated edges are where there are directed edges with opposite orientations or where one CPDAG has a directed edge and the other an undirected one. The values are all scaled by dividing by the number of edges in the true CPDAG. The F1 of the learnt CPDAG is also shown.}
    \label{fig:ord_hc_opt_edges}
\end{figure}

\begin{figure}[htp]
    \centering
    \includegraphics[width=13cm]{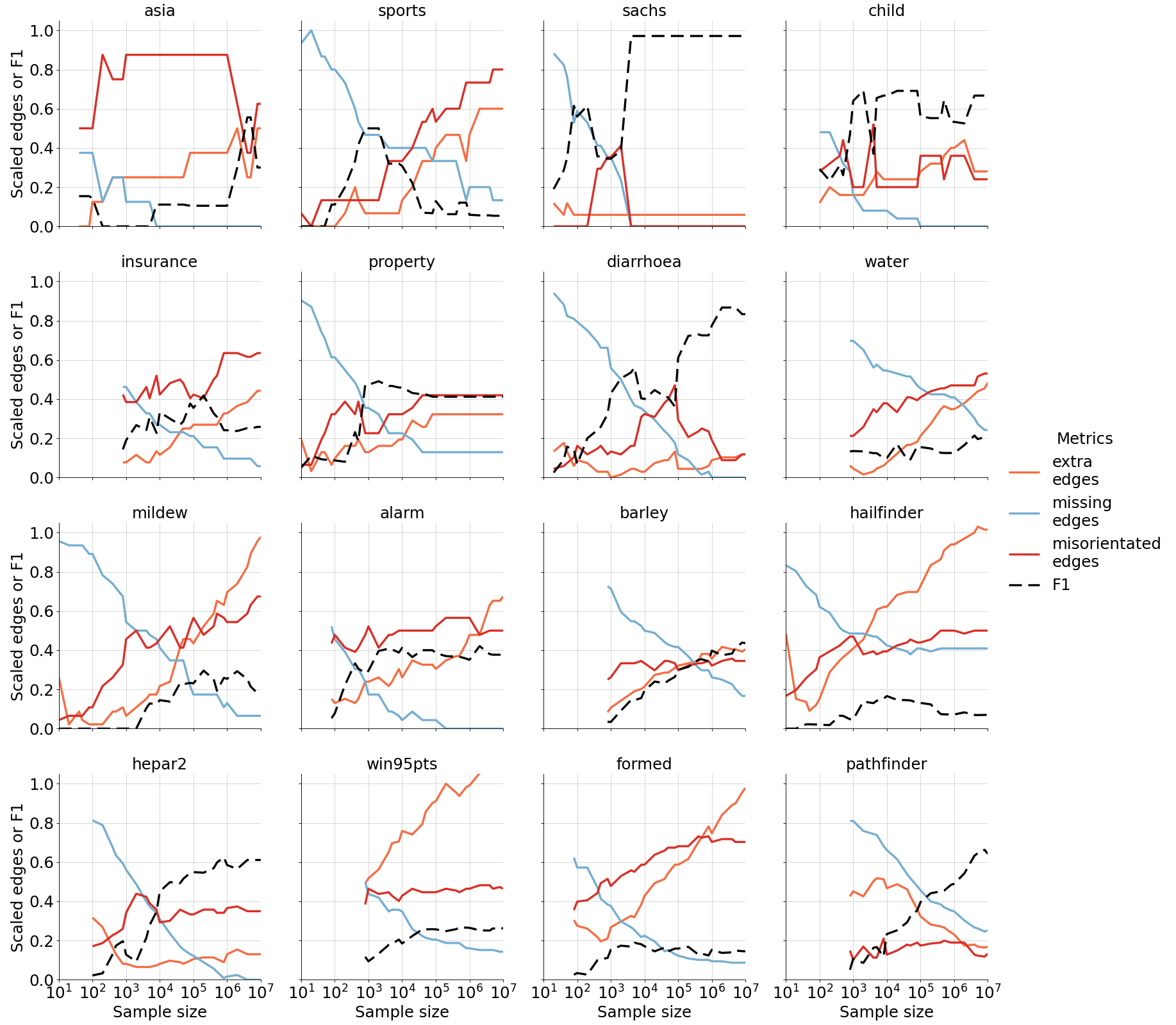}
    \caption{Scaled number of incorrect edges and F1 of learnt CPDAG when \textbf{worst ordering} is used. The red line shows extra edges, either directed or undirected, the blue line shows missing edges and the orange line shows edges which are misorientated. Misorientated edges are where there are directed edges with opposite orientations or where one CPDAG has a directed edge and the other an undirected one. The values are all scaled by dividing by the number of edges in the true CPDAG. The F1 of the learnt CPDAG is also shown.}
    \label{fig:ord_hc_worst_edges}
\end{figure}

\begin{figure}[htp]
    \centering
    \includegraphics[width=13cm]{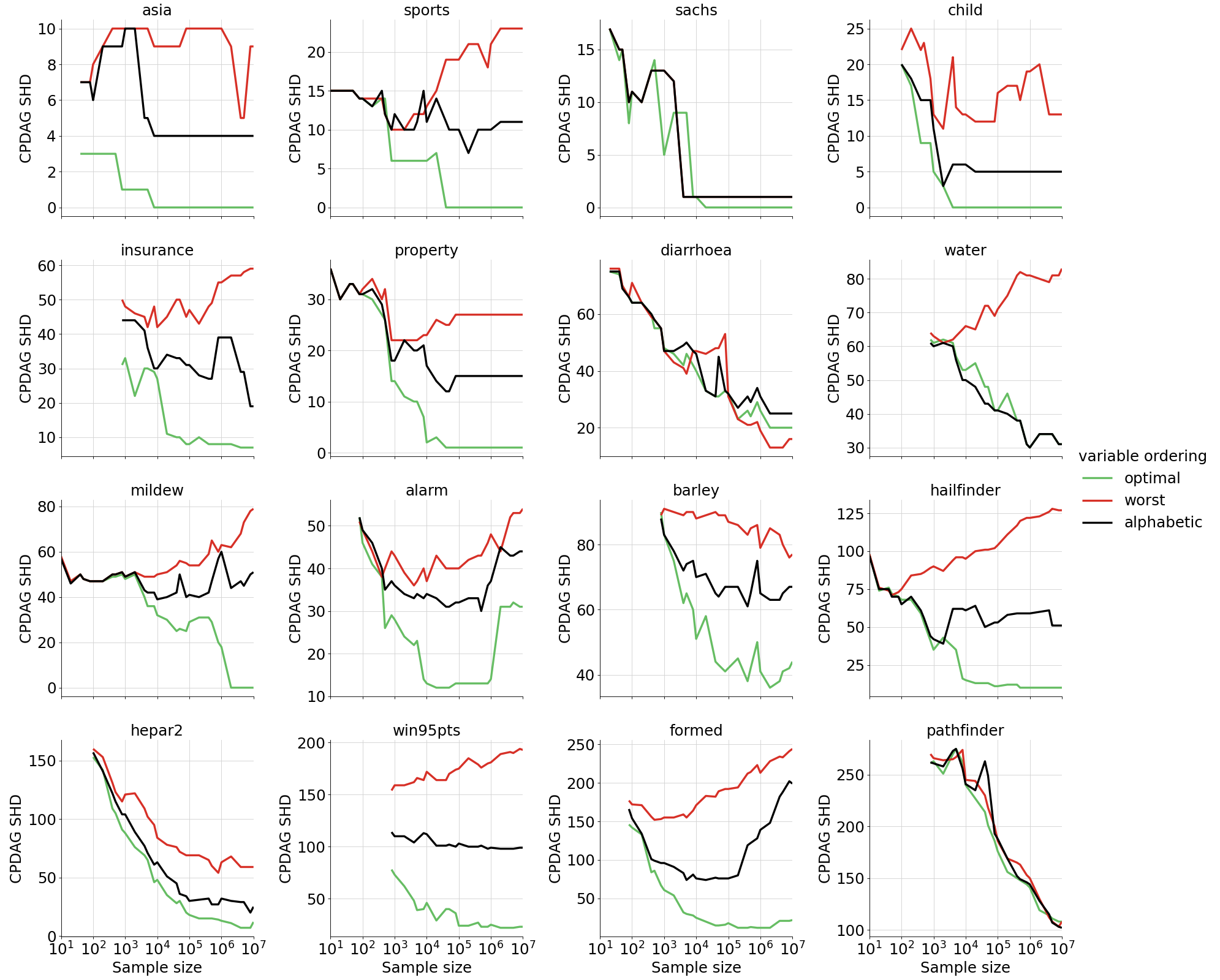}
    \caption{CPDAG SHD against sample size for each variable ordering and network for the HC algorithm. Each plot starts at the sample size at which there are no single-valued variables.}
    \label{fig:ord_hc_shd}
\end{figure}

\begin{figure}[htp]
    \centering
    \includegraphics[width=13cm]{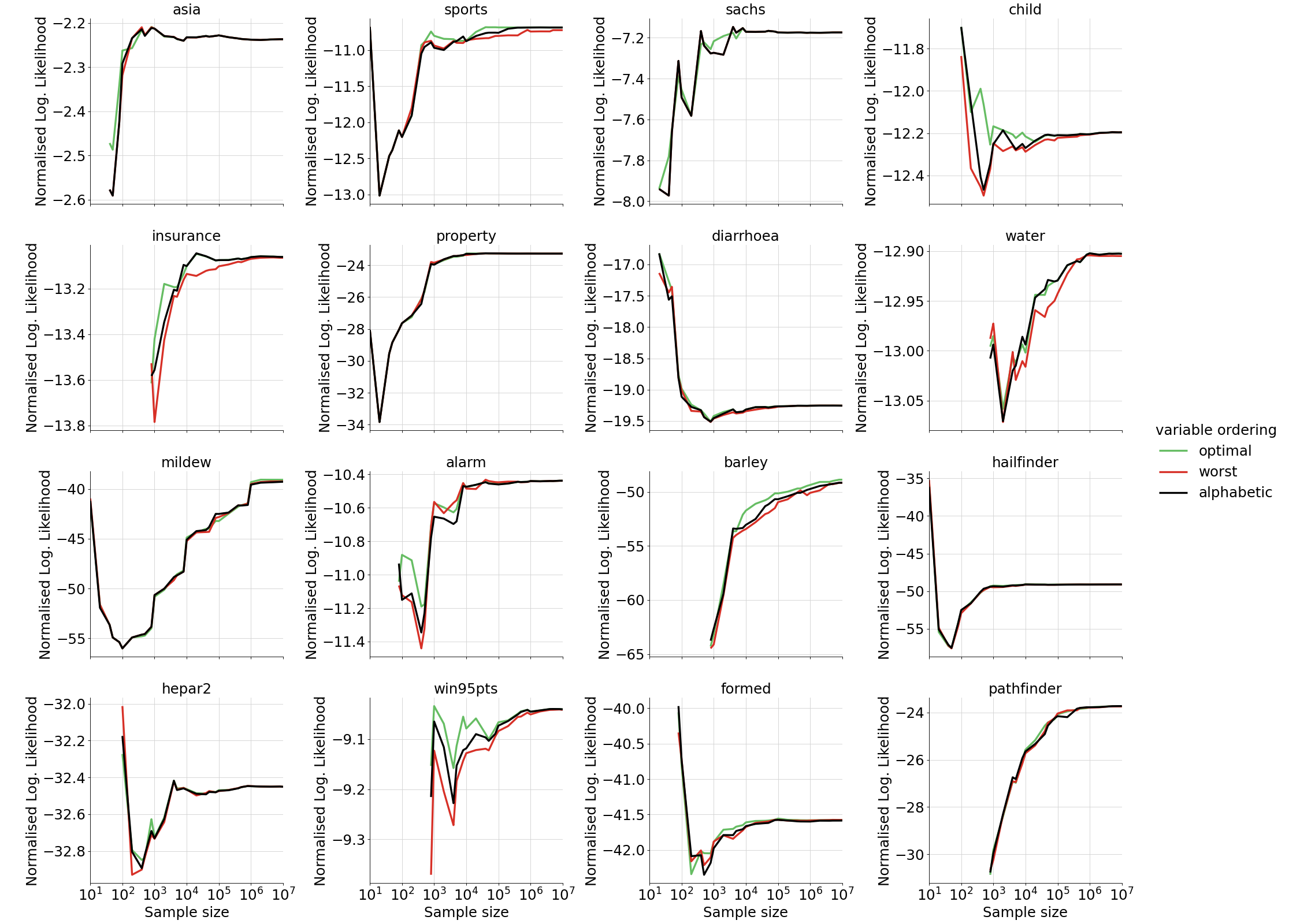}
    \caption{Normalised log-likelihood against sample size for each variable ordering and network for the HC algorithm. Each plot starts at the sample size at which there are no single-valued variables. (Note that the green, red and black lines are drawn in that order, so that where they are all coincident only the black line is visible).}
    \label{fig:ord_hc_loglik}
\end{figure}

\begin{figure}[htp]
    \centering
    \includegraphics[width=13cm]{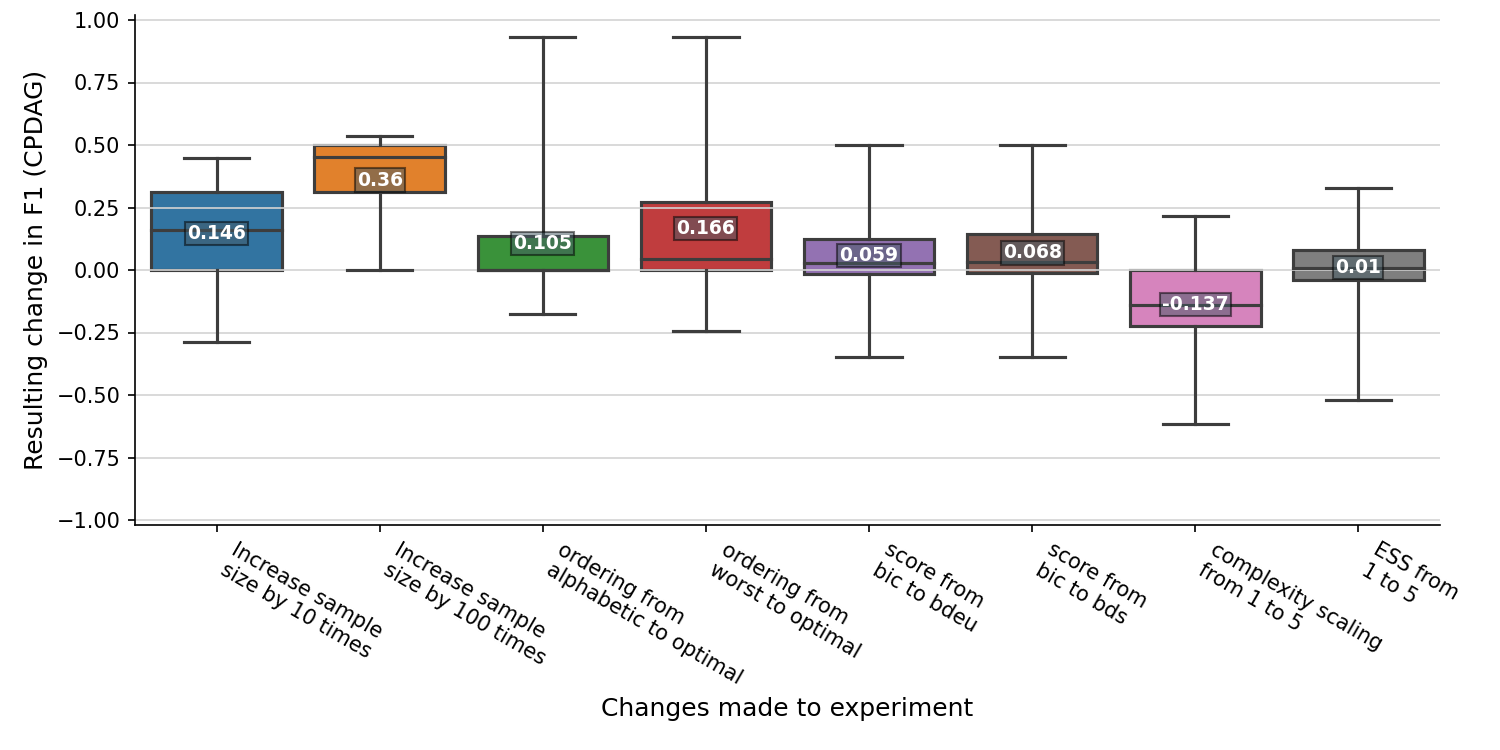}
    \caption{Impact on F1 scores (CPDAG) of changing sample size, variable ordering, score or hyper-parameters across all networks using the HC algorithm with low-dimensional sample sizes between $10$ and $10^3$. Each plot shows the mean change as a number, the median change as a horizontal black line, the interquartile range as the coloured rectangle, and the minimum and maximum values as whiskers.}
    \label{fig:ord_hc_impact_lowd}
\end{figure}

\end{document}